\documentclass[pdflatex,sn-mathphys-num]{sn-jnl}

\usepackage{graphicx}%
\usepackage{multirow}%
\usepackage{amsmath,amssymb,amsfonts}%
\usepackage{amsthm}%
\usepackage{mathrsfs}%
\usepackage[title]{appendix}%
\usepackage{xcolor}%
\usepackage{textcomp}%
\usepackage{manyfoot}%
\usepackage{booktabs}%
\usepackage{algorithm}%
\usepackage{algorithmicx}%
\usepackage{algpseudocode}%
\usepackage{listings}%
\usepackage{subcaption}
\usepackage{float}

\definecolor{lavender}{rgb}{0.9, 0.9, 0.98}
\usepackage{multirow}

\definecolor{lightgray}{gray}{0.92}
\definecolor{headergray}{gray}{0.85}
\usepackage{booktabs}
\usepackage{xcolor,colortbl}

\theoremstyle{thmstyleone}%
\theoremstyle{thmstyletwo}%

\theoremstyle{thmstylethree}%

\begin{document}

\title[Article Title]{Automatic Uncertainty-Aware Synthetic Data Bootstrapping for Historical Map Segmentation}


\author*[1]{\fnm{Lukas} \sur{Arzoumanidis}}\email{lukas.arzoumanidis@hcu-hamburg.de}

\author[2]{\fnm{Julius} \sur{Knechtel}}\email{knechtel@igg.uni-bonn.de}

\author[2]{\fnm{Jan-Henrik} \sur{Haunert}}\email{haunert@igg.uni-bonn.de}
\author[1]{\fnm{Youness} \sur{Dehbi}}\email{youness.dehbi@hcu-hamburg.de}

\affil*[1]{\orgdiv{Computational Methods Lab}, \orgname{HafenCity University}, \orgaddress{\street{Henning-Voscherau-Platz 1}, \city{Hamburg}, \postcode{20457}, \country{Germany}}}

\affil[2]{\orgdiv{Institute of Geodesy and Geoinformation}, \orgname{University of Bonn}, \orgaddress{\street{Meckenheimer Allee 172}, \city{Bonn}, \postcode{53115}, \state{North Rhine–Westphalia}, \country{Germany}}}


\abstract{The automated analysis of historical documents, particularly maps, has drastically benefited from advances in deep learning and its success across various computer vision applications. However, most deep learning-based methods heavily rely on large amounts of annotated training data, which are typically unavailable for historical maps, especially for those belonging to specific, homogeneous cartographic domains, also known as corpora. Creating high-quality training data suitable for machine learning often takes a significant amount of time and involves extensive manual effort. While synthetic training data can alleviate the scarcity of real-world samples, it often lacks the affinity (realism) and diversity (variation) necessary for effective learning. By transferring the cartographic style of a historical map corpus onto modern vector data, we bootstrap an effectively \textit{unlimited} number of synthetic historical maps suitable for tasks such as land-cover interpretation of a homogeneous historical map corpus. We propose an automatic deep generative approach and an alternative manual stochastic degradation technique to emulate the visual uncertainty and noise, also known as aleatoric uncertainty, commonly observed in historical map scans. To quantitatively evaluate the effectiveness and applicability of our approach, the bootstrapped training datasets were employed for domain-adaptive semantic segmentation on a homogeneous map corpus using a Self-Constructing Graph Convolutional Network, enabling a comprehensive assessment of the impact of our data bootstrapping methods. 
}






\maketitle

\section{Introduction}\label{sec1}
Historical maps constitute an essential resource for the reconstruction and monitoring of environmental and spatio-temporal changes, representing a fundamental source of geographic information~\citep{histmaps:scientific_studies}. They provide a detailed depiction of the spatial configuration of the Earth's surface over extended temporal intervals preceding the advent of modern Earth observation technologies. To ensure long-term preservation and accessibility, historical maps are typically digitized and stored as raster images in archival databases. However, this process strips away their inherent semantic content, rendering the underlying spatio-temporal knowledge inaccessible to computational analysis. Consequently, to extract this embedded information in a machine-readable and processable form, historical maps must first undergo semantic interpretation.
\\\\
For many decades, the extraction of semantic information from historical maps was conducted manually or through semi-automated workflows -- processes that were both time-consuming and labor-intensive~\citep{histmaps:scientific_studies}. In recent years, deep neural networks have gained significant traction and demonstrated state-of-the-art performance across numerous tasks related to the automatic analysis of historical maps, including object extraction~\citep{li:synthetic_mapLabels, wu:extractHydrofeatures} and semantic segmentation~\citep{hurni:spatioTemporal, arzoumanidis:scgcn_2025}. Such advancements have facilitated progress in multiple research areas, including road extraction~\citep{UHL:roadextraction, sertel:roadextraction_segtransformer, jiao:roadVector}, urbanization studies~\citep{uhl:building_blocks}, and land-use change detection~\citep{segment_centuries, Mayra2023:land_cover}. Despite these successes, deep learning-based approaches for supervised tasks, such as semantic segmentation, demand substantial quantities of pre-annotated high quality training data -- resources that remain extremely limited for historical maps, especially for single map types or periods~\citep{histmaps:scientific_studies, arzoumanidis:deepgeneration}. 
\\\\
As a result, existing methodologies either rely heavily on manually annotated datasets which is a costly and labor-intensive process that contradicts the very goal of automation or employ transfer learning techniques, such as domain adaptation \citep{wu:domainAdaptation}, by utilizing data from other visual domains. More recently, several emerging approaches have sought to overcome the limitations associated with data scarcity and domain heterogeneity. These include the use of Visual Language Models (VLMs)~\citep{ xia2025mapsam2adaptingsam2automatic, sterzinger2025fewshotsegment}, Large Language Models (LLMs)~\citep{yuan2025leveragingllm, duan2025}, synthetic data generation for domain-specific training~\citep{duan2025, segment_centuries}, and transformer-based architectures for robust and scalable map interpretation~\citep{hurni:spatioTemporal, sertel:roadextraction_segtransformer}. Nevertheless, these approaches often encounter challenges related to domain mismatch and limited generalization, primarily caused by the stylistic heterogeneity of historical map corpora. Such variability arises from differences in typeface, symbology, color palette, and scale~\citep{arzoumanidis:scgcn_2025}, as well as from data-dependent uncertainty, also referred to as aleatoric uncertainty, introduced by map degradation effects such as color fading, folding marks, dust accumulation, or mildew stains.
\\\\
A commonly adopted strategy to overcome the scarcity of annotated data is the automatic generation of synthetic training datasets through simulation. In various domains such as autonomous driving and precision agriculture, synthetic data have been successfully employed for tasks including classification \citep{genai_classif}, object detection \citep{li:synthetic_mapLabels}, and semantic segmentation \citep{Kloukiniotis_2022_CVPR}. Simulated datasets can capture a significant portion of the aleatoric uncertainty inherent to real-world distributions, enabling the creation of large-scale, domain-representative training corpora. This capability renders data-intensive machine learning methods both feasible and effective. Despite its success in other fields, the use of simulation for synthetic data generation in the context of historical maps remains limited. Existing efforts have primarily focused on specific applications such as automated text detection~\citep{li:synthetic_mapLabels, kirsanova2025}, road extraction~\citep{jiao:roadVector} or for the analysis of land-use and land-cover change~\citep{segment_centuries}. 
\begin{figure*}[ht!]
  \includegraphics[width=\textwidth]{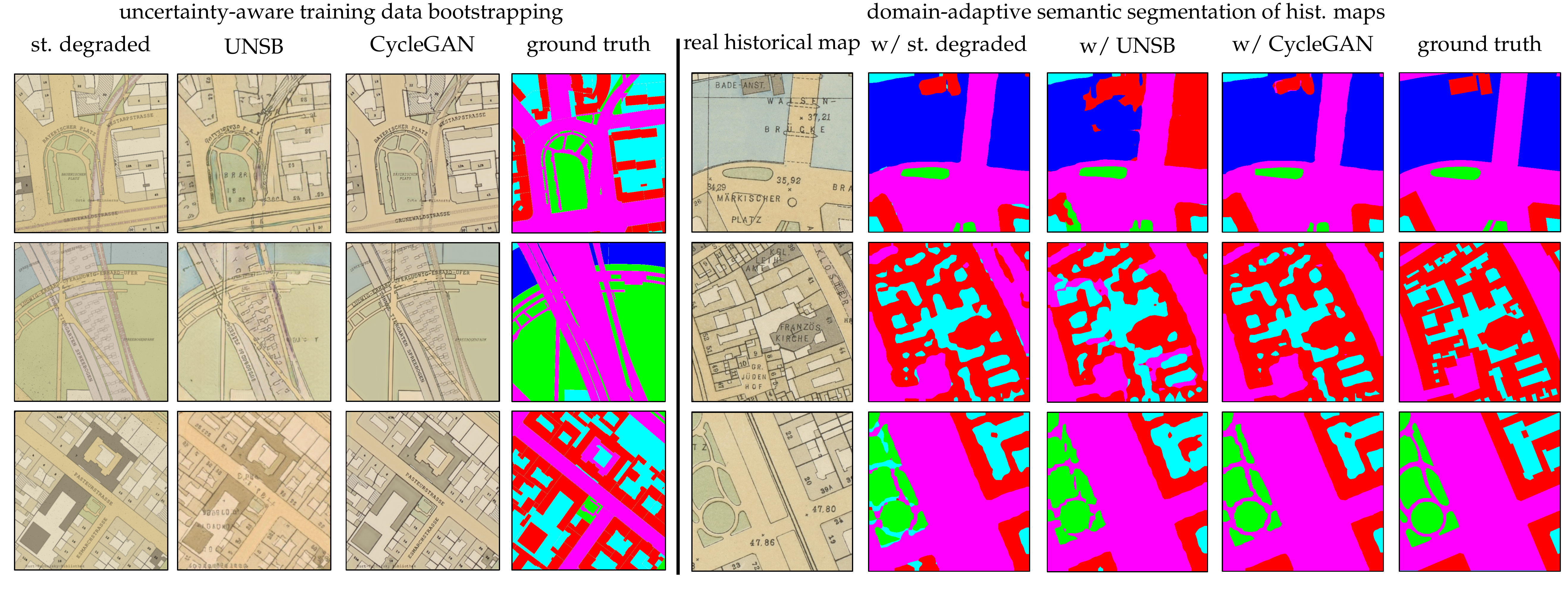}
  \caption{Generated synthetic historical maps with different generation options, such as stochastically (st.) degraded or using CycleGAN and Unpaired Neural Schrödinger Bridge (UNSB) (left). Results of \textit{land-cover prediction} of unseen \textit{real historical maps} using a graph-based deep learning model trained on these synthetic historical maps (right). The learned classes comprise water (blue), sealed surfaces (turquoise), buildings (red), infrastructure (magenta), and recreational surfaces (green).}
  \label{fig:teaser}
\end{figure*}
We propose a novel and highly effective approach for synthesizing historical maps within homogeneous cartographic corpora by employing 
Generative Adversarial Networks (GANs) and Stable Diffusion (SD) --- and benchmarking their performance against stochastic degradation-based image augmentation techniques. Our approach provides explicit control over spatial layouts through a semantically guided, unpaired image-to-image style transfer framework. We validate the effectiveness and practical applicability of our approach by manually annotating historical maps from a homogeneous urban map corpus and performing domain-adaptive semantic segmentation using a model trained exclusively on the bootstrapped historical maps to extract land-cover information, as illustrated in Figure~\ref{fig:teaser}. Explicit control over the spatial layout of semantic vector data is achieved by synthesizing modern, arbitrary vector inputs in historical cartographic styles, enabling the generation of an \textit{unlimited} amount of training data. In this work, we use the term \textit{bootstrapping} to describe the automatic generation of synthetic training data for historical maps from existing modern spatial vector data, without relying on paired training data or manually annotated map examples. To evaluate the practical use of the proposed synthetic data, we trained a state-of-the-art graph-based deep learning model for semantic segmentation. The model, trained exclusively on data bootstrapped by our framework, was subsequently applied to the interpretation of an unseen historical map corpus. To further demonstrate the usability and potential applications of our approach, we developed an interactive web application that showcases its functionality for spatial analyses, such as urban change detection. The source code associated with this work is publicly available at \url{https://github.com/hcu-cml/bootstrap-HistoricalMap}.
\\\\
The main contributions of this work are summarized as follows:
\begin{itemize}
    \item We transfer the cartographic style of a historical map corpus onto arbitrary \textit{modern vector data}, while simulating aleatoric uncertainty to generate \textit{unlimited} annotated training data for a specific homogeneous corpus, 
    \item we evaluate the suitability of our synthetic training data by the Fréchet Inception Distance (FID) and through training a semantic segmentation model for the land-cover prediction on a historical map corpus,
    \item we significantly reduce the time required for manually annotating training data, cutting the process of interpreting land-cover classes from historical maps down to just a few hours.
\end{itemize}

The remainder of this paper is organized as follows: Section~\ref{sec:related_work} reviews recent advances in the synthetic generation of historical maps. Section~\ref{enum_classes} describes the modern vector data and the historical map corpus used in this study. Section~\ref{meth} gives a detailed description of our proposed methodology. Section~\ref{results} evaluates the quality of the bootstrapped training data and demonstrates the effectiveness of our semantic segmentation approach on historical maps. Finally, Section~\ref{conclu} concludes the paper and outlines directions for future research.

\section{Related Work}\label{sec:related_work}
 
Deep learning-based methodologies for the extraction of road networks from historical maps typically require extensive annotated training datasets. To address the scarcity of such data, \citet{jiao:roadVector, JIAO2022102980} and \citet{muehlematter2024roads} introduced a simulation-based framework for road extraction and classification. Their method synthesizes training data by augmenting existing historical road geometries onto historical map backgrounds from the same cartographic corpus. While this strategy effectively expands the amount of available training data, it relies on the assumption that the vector data precisely correspond to the spatial layout of the historical map. In practice, however, this condition rarely holds, as the geometry and topology of road networks often evolve substantially over time. Moreover, the approach requires both the vector and raster datasets to be georeferenced, constraints that are not imposed by our proposed method. 
\\\\
In a related line of research, \citet{affolter_generative_2025} investigated the generation of historical maps with explicit spatial layout control by leveraging Stable Diffusion in combination with ControlNet~\citep{Zhang_2023_ICCV}. Similar to the approaches proposed by \citet{jiao:roadVector, JIAO2022102980} and \citet{muehlematter2024roads}, their method synthesizes historical maps using historical road network vector data as structural input. This form of map bootstrapping has precedents in other domains, such as the creation of two-dimensional figure-ground maps \citep{figure_ground} or the drawing of urban intersection markings in different national styles~\citep{10.1145/3764924.3770902}. However, the approach by \citet{affolter_generative_2025} exhibits certain limitations. Most notably, it depends on the availability of georeferenced historical maps for training and employs a paired image-generation pipeline that integrates ControlNet conditioning with exact historical road network geometries. Such detailed and spatially aligned vector data are rarely available for historical periods, thereby introducing a dependency on pre-existing map interpretations and restricting the applicability of the method to well-documented regions or time frames. In contrast, we propose an approach that is entirely independent of paired training data for the bootstrapping of historical maps by transferring the cartographic style of a historical map corpus onto vector data.
\\\\
Recently, \citet{segment_centuries} introduced a dataset of historical maps designed to support the analysis of large-scale, long-term land-use and land-cover dynamics under limited annotation conditions. The dataset is used to conduct semantic segmentation of land-cover classes with both fully supervised and weakly supervised deep learning models trained on historical labels as well as modern annotations. In the weakly supervised setting, historical maps are first translated into a modern cartographic style using CycleGAN to reduce stylistic discrepancies, thereby enabling segmentation based on existing modern ground-truth labels. In contrast, we propose a domain-adaptive semantic segmentation approach that transfers the cartographic style of historical maps onto \textit{modern vector data}, rather than translating historical maps into a modern style. Furthermore, we aim to reduce stylistic discrepancies in our bootstrapped historical maps by artificially degrading them with aleatoric noise.
\\\\
To advance the automated interpretation of historical maps beyond geometrical feature extraction, the ability to read and understand textual map elements is of increasing importance. Textual labels provide valuable semantic information that supports both map interpretation and metadata enrichment. However, most existing approaches are trained on datasets from unrelated visual domains, resulting in limited generalization to historical cartographic materials. To mitigate this issue, \citet{li:synthetic_mapLabels} proposed a synthetic data generation framework based on cartographic style transfer using GANs. Their model produces synthetic maps in which the placement, typography, and content of text labels can be fully controlled, enabling the creation of large-scale annotated datasets for training text recognition models. Extending this concept, \citet{arzoumanidis:deepgeneration} developed a generative framework that learns the stylistic characteristics of diverse historical map corpora and applies them to contemporary urban maps, thereby producing a virtually unlimited number of synthetic historical maps. Building upon these prior works, our approach directly transfers the learned cartographic style of a historical map corpus onto arbitrary \textit{modern vector data}, generating an effectively unlimited set of synthetic maps with perfectly aligned ground-truth annotations. To the best of our knowledge, such direct coupling between vector-based content and learned cartographic style has not been previously investigated.
\\\\
Complementary to data simulation strategies, several studies have focused on improving model performance under conditions of limited annotated data. Recent research has explored learning paradigms that enhance data efficiency in semantic segmentation and object detection tasks. These include spatio-temporal learning techniques that exploit temporal continuity to reduce annotation requirements \citep{hurni:spatioTemporal, Yuan22102025}, domain adaptation frameworks that enable the transfer of learned representations across heterogeneous map collections \citep{wu:domainAdaptation}, and unsupervised or self-supervised learning approaches that minimize reliance on labeled samples \citep{xia2025mapsam2adaptingsam2automatic, xia2024mapsamadaptingsegmentmodel}. 
\\\\
Another emerging direction leverages the power of pre-trained foundation models and large multimodal networks for cartographic analysis \cite{xia2025mapsam2adaptingsam2automatic, xia2024mapsamadaptingsegmentmodel}. For instance, \citet{yuan2025smolmapseglabel} and \citet{sterzinger2025fewshotsegment} demonstrated that few-shot segmentation models can be effectively adapted to historical maps, while \citet{yuan2025leveragingllm} and \citet{Wang26052025} explored the integration of LLMs to support multimodal reasoning and semantic labeling. Collectively, these approaches highlight a paradigm shift from data-intensive learning toward methods that exploit prior knowledge, transfer learning, and generative modeling to mitigate data scarcity in the analysis of historical maps.

\section{Map Corpus \& Modern Vector Data}\label{enum_classes}
\begin{figure}[ht!]
\centering
\begin{subfigure}{.47\columnwidth}
  \includegraphics[width=\linewidth]{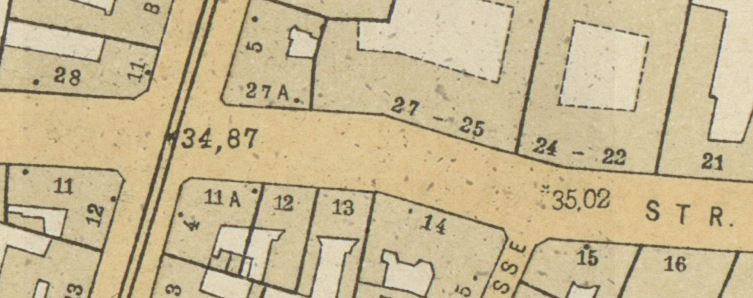}
\end{subfigure}
\hfill
\begin{subfigure}{.495\columnwidth}
  \includegraphics[width=\linewidth]{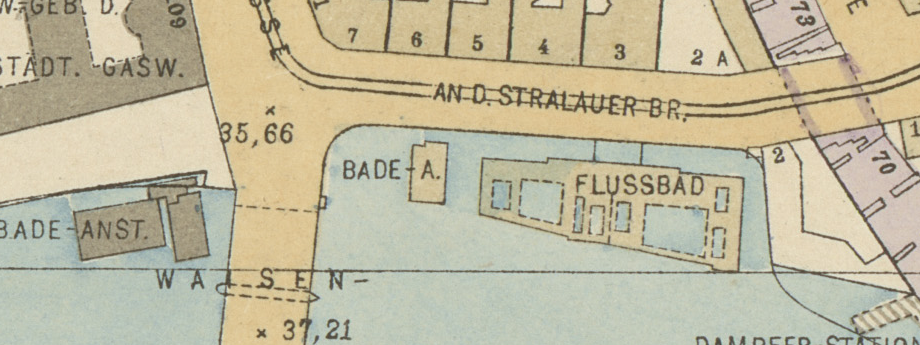}
\end{subfigure}
\caption{Two exemplary map patches of the Straube map corpus. Accumulations of dust and mildew stains are visible, along with imprecisions in shading and coloring.}
\label{fig:noise-uncertrainty}
\end{figure}

The map corpus exemplarily worked with in this study comprises $39$ historical urban maps of Berlin, Germany, provided by the State Library of Berlin, and represents five distinct land-cover classes, namely: 
\begin{enumerate}
    \item \textbf{Buildings} -- including public, private, military, and religious structures;
    \item \textbf{Infrastructure} -- such as bridges, tunnels, streets, and railway tracks;
    \item \textbf{Recreational surfaces} -- including parks, gardens, and cemeteries;
    \item \textbf{Sealed surfaces} -- such as arterial roads, squares and backyards;
    \item \textbf{Water bodies} -- comprising rivers, ponds, and fountains.
\end{enumerate}
All maps of this corpus were produced in the late 19\textsuperscript{th} and early 20\textsuperscript{th} century by \textit{Julius Straube}, a German cartographer. This map corpus is referred to as \textit{Straube maps} throughout the remainder of this work. Figure~\ref{fig:noise-uncertrainty} illustrates two exemplary patches of two different Straube maps contained in the corpus. The maps share a common scale of 1:4000 and a consistent cartographic style, which allows them to be characterized as a homogeneous corpus. Since the primary objective of this work is the semantic segmentation of land-cover classes, we disregard the map frame entirely and focus exclusively on the map content itself. The \textit{modern vector data} used in this study were primarily sourced from OpenStreetMap (OSM)~\citep{OpenStreetMap}. This dataset provides detailed, real-world geometries for building footprints, land parcels, and street networks, making it well suited for cartographic reproduction. To achieve a more faithful replication, this foundational data were enriched with semantic information from Germany’s Authoritative Cadastral Information System (ALKIS)\footnote{\url{https://www.berlin.de/sen/sbw/stadtdaten/geoportal/liegenschaftskataster/alkis/}}(German: \textit{Amtliches Liegenschaftskatasterinformationssystem}, ALKIS). 

\section{Methods}\label{meth}
\begin{figure}[ht!]
  \centering
  \includegraphics[width=\columnwidth]{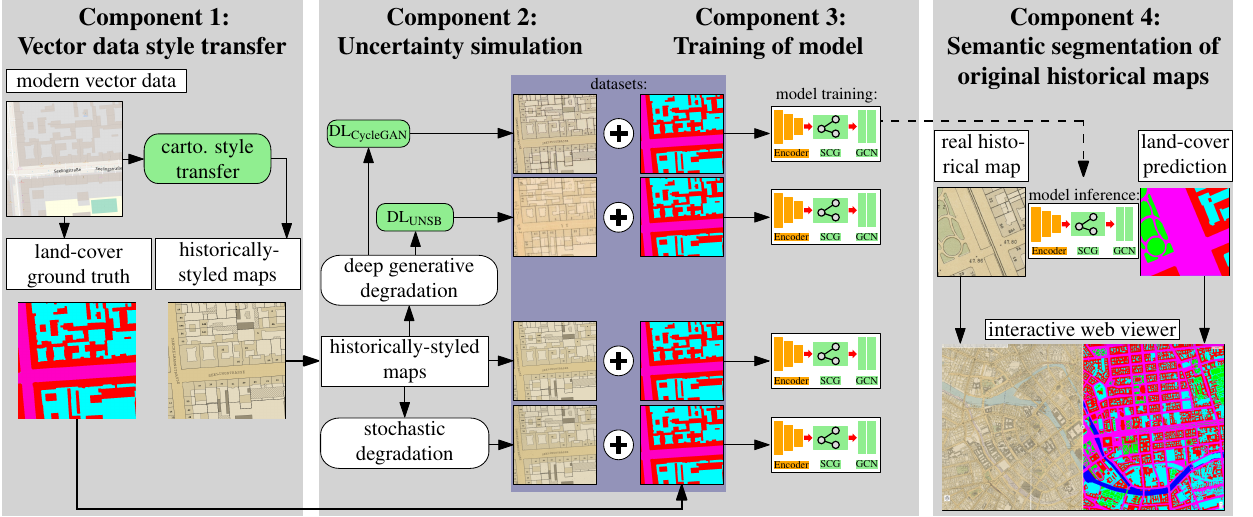}
  \caption{
  By applying cartographic (carto.) style transfer to the color scheme and line thickness of \textit{modern vector data}, we generate \textit{historically styled maps}, whereas the \textit{land-cover ground truth} is derived directly from the semantic information of the \textit{modern vector data} (1). Aleatoric uncertainty is simulated through either deep generative or stochastic degradation of the \textit{historically-styled maps} (2). By combining the degraded \textit{historically-styled maps} with their \textit{land-cover ground truth} we obtain four different bootstrapped datasets. We select one bootstrapped dataset and perform domain-adaptive learning with a graph-based neural network (3). Finally, the semantic segmentation of an unseen historical map corpus (Straube Maps, cf. Section~\ref{enum_classes}) is performed using the trained graph-based deep learning model (4).}
  \label{method_pipeline}
\end{figure}

In this section, we describe how \textit{modern vector data}, can be utilized to generate an \textit{unlimited} number of training samples for domain-adaptive semantic segmentation of historical maps by explicitly controlling both the spatial composition and the semantic layout of the synthesized maps. The proposed workflow comprises four main components. First, we present our method for generating realistic synthetic \textit{historically-styled maps} that replicate the cartographic style of a specific homogeneous map corpus. We also describe how the corresponding \textit{land-cover ground truth} are generated to produce paired training samples required for deep learning-based semantic segmentation (cf. Figure~\ref{method_pipeline}, Component~1). 
\\\\
Secondly, we detail our approach for simulating aleatoric uncertainty by augmenting the \textit{historically-styled maps} with degradation effects derived from deep generative models or, alternatively, from manual stochastic degradation techniques (cf. Figure~\ref{method_pipeline}, Component~2). To automatically degrade the \textit{historically-styled maps} with simulated aleatoric uncertainty, we employ two different deep learning-based generative approaches, namely, \textbf{Generative Adversarial Networks} and \textbf{Stable Diffusion}, employing unpaired image-to-image translation. At this stage, we obtain four different types of synthetic historical map datasets: \textit{historically-styled maps}, stochastically degraded \textit{historically-styled maps}, deep generative degraded \textit{historically-styled maps} bootstrapped using GAN, deep generative degraded \textit{historically-styled maps} bootstrapped using SD.
\\\\
Finally, we demonstrate how the four resulting synthetic historical map datasets are employed to train a graph-based deep learning framework, yielding four differently trained models (cf. Figure~\ref{method_pipeline}, Component~3). One of these models is subsequently selected to perform domain-adaptive semantic segmentation on previously unseen \textit{real historical maps}, with the results visualized through an interactive web-based viewer (cf. Figure~\ref{method_pipeline}, Component~4). 
\\\\
More formally, our goal is to learn a mapping $G: X \rightarrow Y$ such that the translation $G(x)$ of any generated map $x \in X$ simulates the aleatoric uncertainty within the domain of a historical map corpus $Y$, while preserving the cartographic style and semantic content, such as the spatial layout of the \textit{modern vector data} underlying $x$.
\\\\
Aleatoric uncertainty arising from the physical deterioration of map sheets, archival storage conditions, or suboptimal scanning processes further complicates the generation of suitable training data for deep learning-based approaches. To ensure realistic learning behavior, the generated synthetic historical map datasets must accurately represent these inherent uncertainties across the entire data distribution. Typical degradation effects include paper distortion, folding marks, and additional impurities or artifacts such as dirt, dust, mildew stains, and color fading (cf. Figure~\ref{fig:noise-uncertrainty}).

\subsection{Modern Vector Data Style Transfer}

\begin{figure}[ht!]
\centering
\begin{subfigure}{.24\columnwidth}
  \centering
        \includegraphics[width=\linewidth]{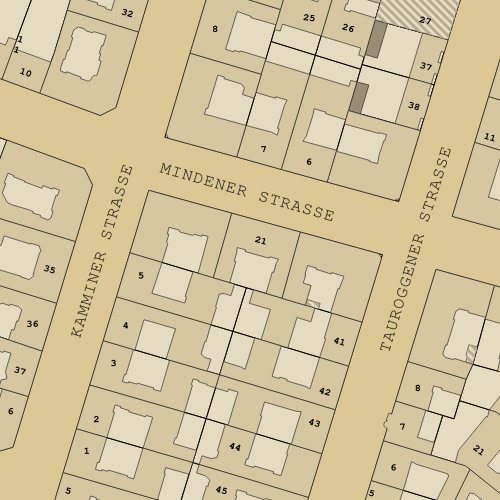}
        \caption{historically-styled historical map}
        \label{fig:vector_synth_map}
\end{subfigure}
\hfill
\begin{subfigure}{.24\columnwidth}
  \centering
        \includegraphics[width=\linewidth]{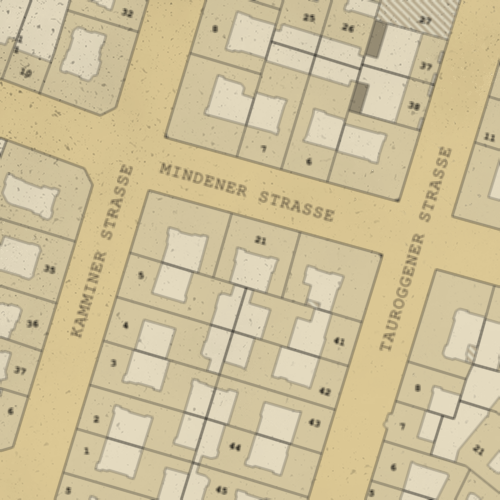}
        \caption{stochastically degraded map}
        \label{fig:enhanced_M}
\end{subfigure}
\hfill
\begin{subfigure}{.24\columnwidth}
  \centering
        \includegraphics[width=\linewidth]{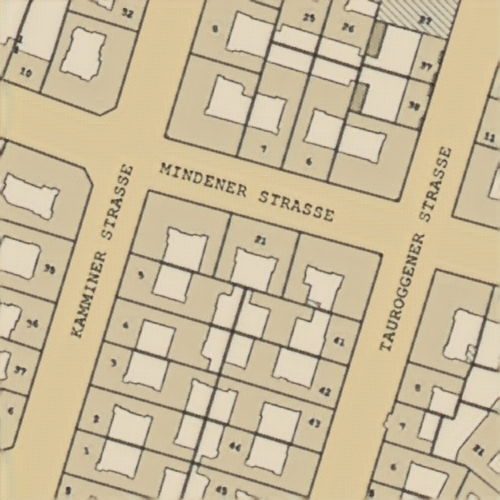}
    \caption{DL\textsubscript{CycleGAN}-based degraded map}
        \label{fig:enhanced_DL}
\end{subfigure}
\hfill
\begin{subfigure}{.24\columnwidth}
  \centering
        \includegraphics[width=\linewidth]{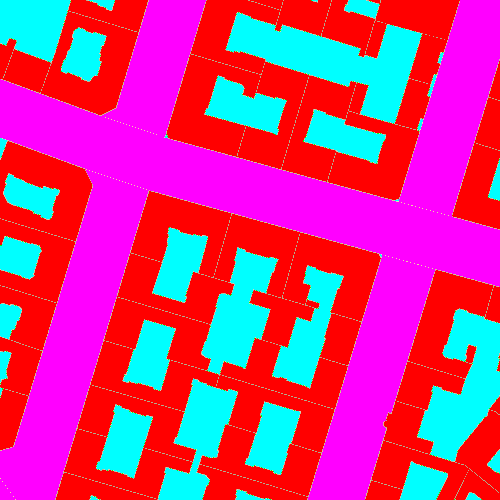}
        \caption{land-cover ground truth}
        \label{fig:vector_styled_annotation}
\end{subfigure}
    \caption{Bootstrapped map examples comprising a purely \textit{historically-styled map} (a), a stochastically degraded \textit{historically-styled map} (b), a \textit{historically-styled map} degraded with simulated aleatoric uncertainty using DL\textsubscript{CycleGAN} (c) and the corresponding \textit{land-cover ground truth} representing the semantic classes (d). Turquoise denotes sealed surfaces, red denotes buildings and magenta denotes infrastructure.}
    \label{fig:vector_styled_map_and_annotation}
\end{figure}

Visual inspection, with limited manual effort, is employed to reproduce semantic map objects, including parcel and building boundaries, streets, water bodies, and green spaces, as well as the font types and sizes of a historical map corpus. On this basis, the cartographic style of individual \textit{real historical maps} within the corpus is extracted. To mitigate the influence of impurities, color fading, and compression artifacts when estimating the color or line thicknesses for each map object and assigning it to a \textit{land-cover ground truth} class, we computed the mean RGB value within a 5×5 kernel. Given that the \textit{modern vector data} are already semantically categorized, the color representation of individual classes can be readily adjusted using visualization tools such as  MapTiler\footnote{\url{https://www.maptiler.com}} or Maputnik\footnote{\url{https://maputnik.github.io}}, as demonstrated in our interactive web-based viewer\footnote{\url{https://www.cml.hcu-hamburg.de/demos/historicalMaps-bootstrapping}}. In addition to the aforementioned elements, a coordinate grid was overlaid onto the \textit{modern vector data} to replicate the layout conventions observed in the historical map corpus used in this study. Furthermore, the \textit{historically-styled maps} are semantically enriched with house numbers within building footprints and street names aligned with their corresponding street geometries. Besides, we overlay the names of cultural, historical, medical, and public institutions, as well as notable places, monuments, and landmarks, utilizing the semantic information derived from the \textit{modern vector data}. The total time required to reproduce the cartographic style of a historical map corpus, such as the Staube maps, is less than two hours. 
\\\\
To generate map pairs consisting of a \textit{historically-styled map} and its corresponding \textit{land-cover ground truth}, the \textit{modern vector data} were duplicated and aggregated according to the land-cover classes defined in Section~\ref{enum_classes}. Notably, our approach allows users to specify custom map dimensions, enabling the generation of training samples tailored to all kinds of homogeneous cartographic domains of varying and irregular formats. Figure~\ref{fig:vector_styled_map_and_annotation} explicitly illustrates this bootstrapping process by showing (a) a purely \textit{historically-styled map}, (b) a stochastically degraded \textit{historically-styled map}, (c) a \textit{historically-styled map} with simulated aleatoric uncertainty generated using $DL\textsubscript{CycleGAN}$, and (d) the corresponding ground-truth semantic mask. The training data bootstrapped through this process are referred to as the \textbf{historically-styled} dataset.

\subsection{Aleatoric Uncertainty Simulation}\label{data_uncertainty}

\begin{figure}[ht!]
  \centering
\centering
\begin{subfigure}{.565\columnwidth}
  \includegraphics[width=\linewidth]{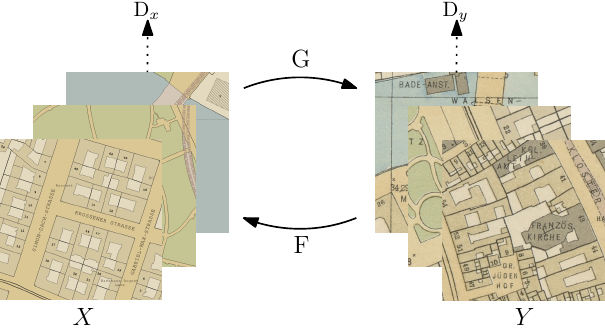}
  \caption{Generator-Discriminator interplay}
  \label{fig:cycleGANleft}
\end{subfigure}
\hfill
\begin{subfigure}{.42\columnwidth}
  \includegraphics[width=\linewidth]{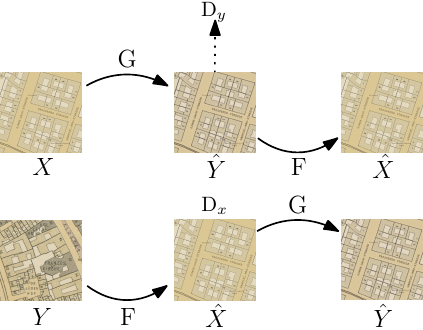}
  \caption{cycle-consistency process on maps}
  \label{cycleGANright}
\end{subfigure}
  \caption{Generator-Discriminator architecture of the underlying CycleGAN for aleatoric uncertainty simulation in historically-styled Straube maps.}
  \label{architectureGAN}
  \end{figure}

By simulating aleatoric uncertainty to enhance the realism of the generated \textit{historically-styled maps} and provide a more robust foundation for domain-adaptive semantic segmentation, we introduce a stochastic degradation process applied to the \textit{historically-styled maps}, as illustrated in Figure~\ref{method_pipeline} (Component 2). Since historical maps often exhibit perceptually complex and noisy backgrounds caused by imperfections introduced during their preservation, the generated \textit{historically-styled maps} were degraded to more faithfully replicate the cartographic style, including its inherent uncertainties, of a \textit{real historical map} corpus by simulating aleatoric uncertainty. Although such uncertainty cannot be effectively modeled directly \citep{hurni:spatioTemporal, hurni:scanned_uncertainty}, we instead simulate aleatoric uncertainty within the generated \textit{historically-styled maps} themselves.
\\\\
\textbf{Stochastic degradation}. To this end, we first apply a Gaussian filter with a kernel size of $3 \times 3$ to the \textit{historically-styled maps}. This operation smooths sharp edges and enhances visual realism, thereby approximating the aleatoric uncertainty typically observed after scanning historical map sheets. In addition, we introduce further uncertainty by randomly cropping, rotating, and overlaying a semi-transparent image of black dust\footnote{\url{https://pngtree.com/freepng/dust-overlay-element-png_8071784.html}} onto the \textit{historically-styled maps}. This step is intended to mimic common visual artifacts found in historical maps, such as dust accumulation and mildew stains (cf. Figure~\ref{fig:noise-uncertrainty}). An example training pair comprising a synthetic \textit{historically-styled map} degraded through manual simulation of such aleatoric uncertainty is shown in Figure~\ref{fig:enhanced_M}. Consequently, we call this second synthetically bootstrapped training dataset \textbf{stochastically (st.) degraded}, which incorporates minimal manual effort to improve the purely \textit{historically-styled map} through the simulation of aleatoric uncertainty, as illustrated in Figure~\ref{method_pipeline} (Component 2). When rendering the purely \textit{historically-styled maps} in vector format, anti-aliasing is enabled to reduce visual artifacts and eliminate the jagged appearance, thereby enhancing the overall visual quality of the maps.
\\\\
\textbf{Deep generative degradation}. To further advance our approach and explore methods for enhancing the \textit{historically-styled maps} without manual intervention, we propose the use of deep generative models. Unlike our stochastic degradation method that relies on manual finetuning, these models are supposed to learn the aleatoric uncertainty directly from the underlying \textit{real historical map} corpus enabling fully automated degradation. We selected the cycle-consistent adversarial network CycleGAN~\citep{cycleGAN} for our GAN-based experiments because it is widely recognized for achieving competitive results in image generation across diverse scientific domains and, more importantly, has already been successfully applied to the cartographic domain \citep{arzoumanidis:deepgeneration, li:synthetic_mapLabels}. For our second deep generative approach, we selected Stable Diffusion (SD), as it has recently demonstrated promising results for the generation of historical maps~\citep{affolter_generative_2025}. Specifically, we utilized Unpaired Neural Schrödinger Bridge (UNSB), which has recently been successfully applied to unpaired image-to-image translation tasks and style transfer in complex visual domains~\citep{kim2023unsb}. As a result, we bootstrap a third and forth training dataset, as illustrated in Figure~\ref{method_pipeline} (Component 2). To facilitate the distinction between each training dataset and its corresponding model, we introduce \textbf{DL\textsubscript{CycleGAN}} dataset and \textbf{DL\textsubscript{UNSB}} dataset, which represent the incorporation of deep learning-based, aleatoric uncertainty simulation. 
\\\\
The CycleGAN model learns to translate the style from the input domain $Y$ to the input domain $X$ in an unpaired manner, using an architecture composed of two image Generators, $G$ and $F$, and two image Discriminators, $D_{x}$ and $D_{y}$, as illustrated in Figure~\ref{fig:cycleGANleft}. The architecture of an image Generator is composed of convolutional layers that encode latent features into a higher-dimensional space by progressively increasing the number of image channels. In our implementation, the CycleGAN architecture was adapted to process input images with dimensions of $500 \times 500 \times 3$. Residual blocks are then applied to preserve high-quality image details while maintaining style consistency and semantic fidelity. Finally, a decoder upsamples the high-dimensional feature representation back to the original image resolution of $500 \times 500 \times 3$~\citep{cycleGAN}. The Discriminator architecture is based on a PatchGAN module, which evaluates whether an input image is real or generated. To ensure generative realism in the synthesized historical maps and to enforce cycle-consistency, i.e., the ability to transform \textit{historically-styled maps} into degraded maps and to reconstruct them back, CycleGAN employs two distinct loss functions, each serving a specific purpose. The adversarial loss incentivizes the Generator to produce outputs that are indistinguishable from real images $Y$ within the target domain $X$. To prevent arbitrary permutations of images within the target domain, a cycle-consistency loss is introduced. This loss constrains the space of possible mapping functions by enforcing that, for each image $x$ from domain $X$, the translation cycle should be capable of reconstructing the original image~\citep{cycleGAN}.
\\\\
In our setup, set $X$ corresponds to the bootstrapped historical maps, while set $Y$ comprises maps from a homogeneous historical corpus. Generator $G$ translates maps from domain $X$ into the style of domain $Y$, producing $\hat{Y}$. This translated map is then passed through Generator $F$, which reconstructs it back into domain $X$, yielding $\hat{X}$. In the second cycle, maps from set $Y$ are translated into domain $X$ using Generator $F$, resulting in $\hat{X}$, which are subsequently translated back to domain $Y$ using Generator $G$, producing $\hat{Y}$. This cycle-consistency process is illustrated in Figure~\ref{cycleGANright}.
  \\\\
For generating perceptually realistic historical maps with simulated aleatoric uncertainty, we evaluate an early training epoch, selected before the onset of noticeable changes in map geometry, object topology or semantics. We found that this approach effectively captures early-stage cartographic style refinements and aleatoric uncertainty while avoiding later-stage translations that could compromise control over the spatial layout of the semantic \textit{modern vector data}. To validate early-stage checkpoints, we assess the generated historical maps through \textbf{Fréchet Inception Distance (FID)} scores \citep{paper:fid}. This ensures that the synthetic training data remains both stylistically consistent and topologically faithful to the \textit{land-cover ground truth}.
\\\\
The Unpaired Neural Schrödinger Bridge (UNSB) model formulates unpaired image-to-image translation as a discretised Schrödinger Bridge (SB) problem. Within the SB framework, the objective is to identify the most probable stochastic process $(X_t)_{t \in [0,1]}$ that transports a source distribution $p_0$, representing the cartographic style of historical maps, to a target distribution $p_1$, corresponding to the historically-styled \textit{modern vector data}, while maintaining proximity to a prescribed reference diffusion process $q$. This can be expressed as minimising the Kullback-Leibler divergence. Because solving this optimisation directly is infeasible for high-dimensional images, the Unpaired Neural Schrödinger Bridge (UNSB) approximates the continuous-time solution through a series of discrete steps $X_{t_{k+1}} = G_{\theta_k}(X_{t_k})$, where each generator $G_{\theta_k}$ performs a small, incremental transformation. Instead of learning a single complex mapping, the model constructs a chain of simpler transitions that, together, approximate the SB path from the cartographic appearance of the \textit{real historical maps}, including its aleatoric uncertainty, to the \textit{historically-styled maps}. To guide this progression, UNSB employs a time-conditioned Discriminator that assesses whether intermediate samples resemble the corresponding stage of the target distribution. Although adversarial losses are used, each step essentially receives feedback on how closely its outputs match the cartographic characteristics of the historically-styled \textit{modern vector data}, without requiring explicit likelihood modelling. To preserve geographic structure, UNSB incorporates lightweight identity and consistency regularisers that encourage $G_{\theta_k}(x) \approx x$ for inputs that should maintain their spatial form, similar to the cycle-consistency loss that enforces the ability to map a generated historical map forward and then back to its original form in CycleGAN~\citep{kim2023unsb}. These constraints help prevent distortions of map geometry, such as boundaries, networks, or contours, ensuring that stylistic translation does not compromise the semantic information required for downstream segmentation tasks.
\\\\
While CycleGAN also enables unpaired image-to-image translation, its mechanism differs fundamentally from the UNSB framework. CycleGAN learns two deterministic mappings between domains, constrained by a cycle-consistency loss that enforces the ability to map an image forward and then back to its original form. Although this promotes content preservation, it does not provide a principled way to model the distributional evolution between domains. In contrast, UNSB constructs a stochastic transport path between the cartographic style and aleatoric uncertainty of the \textit{real historical maps} and cartographic appearance of the \textit{historically-styled maps}, decomposing the translation into a sequence of intermediate steps that approximate a Schrödinger Bridge. Rather than relying on cyclic reconstruction to maintain structural alignment, UNSB achieves semantic consistency through gradual transport, distribution matching at each step, and explicit regularisation. This leads to a smoother and more controlled transformation from cartographic style and aleatoric uncertainty of the \textit{real historical maps} to the \textit{historically-styled maps} and helps avoid the abrupt or unstable shifts that can occur in CycleGAN-based mappings, which is the rational behind our early-stopping strategy when training CycleGAN.

\subsection{Domain-Adaptive Semantic Segmentation}

\begin{figure}[ht!]
  \centering
  \includegraphics[width=0.7\columnwidth]{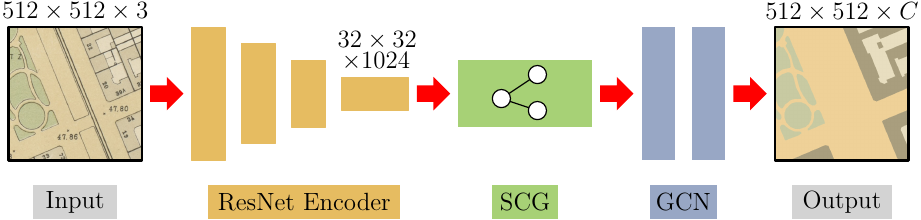}
  \caption{Architecture of the underlying Self-Constructing Graph Convolutional Network (SCGCN), after \citet{arzoumanidis:scgcn_2025}, which contains three parts: ResNet feature encoder, Self-Constructing Graph Module and Graph Convolutional Network.}
  \label{architectureSCGGCN}
\end{figure}

To assess the realism of the synthetically generated maps and their effectiveness in the context of domain-adaptive semantic segmentation, we employ two evaluation strategies. (a) We compute the \textbf{Fréchet Inception Distance (FID)} to quantitatively measure the similarity between the synthetic and \textit{real historical maps}. (b) We train \textbf{Self-Constructing Graph Convolutional Networks (SCGCN)} for land-cover classification and evaluate their performance on $125$ manually annotated samples from the Straube maps, which the synthetically generated maps aim to replicate.
\\\\
The SCGCN employed in this work was originally introduced by \citet{Liu2020_1SCGwACW} and has demonstrated strong performance in various semantic segmentation tasks, including urban scene segmentation in remote sensing imagery \citep{Liu2020_3SCGLongRange}, two-branch segmentation of indoor floorplans \citep{knechtel2024FloorplanSegmentation}, and most importantly in the segmentation of heterogeneous historical map corpora \citep{arzoumanidis:scgcn_2025}. In their study, \citet{arzoumanidis:scgcn_2025} demonstrated that, for heterogeneous historical map corpora, the application of an SCGCN yields improved performance relative to traditional CNN-based approaches. Specifically, they compared their SCGCN with a state-of-the-art CNN-based method built on the U-Net architecture proposed by \citet{petitpierre:cnn} and showed that the SCGCN outperformed the U-Net-based model on a heterogeneous corpus. One explanation is that the SCGCN is more effective at modeling long-range dependencies through a self-constructing graph that enlarges the receptive context, whereas U-Net-based approaches are often inherently local and may therefore fail to capture the broader spatial patterns characteristic of many cartographic structures. Based on these findings, we consider SCGCN as the state-of-the-art model for the semantic segmentation of historical maps and, therefore, select it for our application.
\\\\
For the semantic segmentation of a respective input map, each pixel needs to be assigned to one of the five land-cover classes $C = \{$building, infrastructure, recreational surfaces, water bodies, sealed surfaces$\}$. The key advantage of this architecture lies in its ability to capture long-range dependencies within an image beyond local neighborhoods, without requiring an excessively deep network, by leveraging graph convolutional layers. The graph structure itself is learned automatically, eliminating the need for domain-specific expert knowledge. 
An overview of the SCGCN architecture is shown in Figure~\ref{architectureSCGGCN}.
The network consists of three parts. First, the input image is fed into a pretrained ResNet Encoder, in our case we use three layers of a ResNet-50. Second, the resulting features are used in the Self-constructing graph module (SCG), which forms the core of the network. Here, a graph structure is automatically induced, in which the features directly form the nodes. In addition, edges between two features are added based on their respective similarity. For this process, a new embedding for the features is learned based on Gaussian parameters $\mu$ and $\sigma$. The latent variables are regularized by calculating the Kullback-Leibler divergence between a centered isotropic Gaussian prior distribution over the parameters and the new embedding. By essentially performing a similarity check between each pair of embeddings, edges are added between similar features. Third, the resulting graph is then fed into two graph convolutional layers, in the figure noted as GCN, which assign each pixel of the input image to one of the classes $C$. To overcome the well-known problems associated with imbalanced classes in a dataset, \citet{Liu2020_1SCGwACW} introduced the Adaptive Class Weighting Loss (ACW loss) which we applied in our work as well. For this, weights for the respective classes are computed by performing an iterative batch-wise class rectification instead of, using a fixed weight based on the frequency of occurrence in the whole dataset.

\section{Experimental Results}\label{results}

\begin{figure}[ht!]
  \centering
  \includegraphics[width=.5\columnwidth]{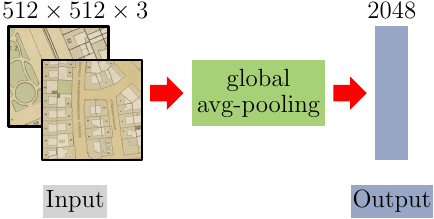}
  \caption{Adapted architecture illustrates how both the Straube maps and the \textit{historically-styled maps} are transformed into a high-level feature space using the final layer of the Inception-v3 network, following the approach of \citet{paper:fid}.}
  \label{architectureInception}
\end{figure}

To create a seamless mosaic of segmentation results, the Straube maps were manually georeferenced. For the sake of reproducibility and to support future benchmarking, the manually annotated test dataset for the Straube maps is publicly available at \url{https://zenodo.org/records/17579452}. The spatial consistency between adjacent image patches, which were processed independently during semantic segmentation, is crucial, as the segmented patches are reassembled into a seamless mosaic to reconstruct the full geographic extent of the Straube map, supporting further analysis of the land-cover classes. For the sake of illustrating our bootstrapped training datasets, the semantic segmentation results and the full geographic extent of the Straube maps, we developed a front end visualization, based on OpenLayers\footnote{\url{https://openlayers.org/}}, hosted as a website at \url{https://www.cml.hcu-hamburg.de/demos/historicalMaps-bootstrapping}. Alongside the \textit{historically-styled maps} generated from \textit{modern vector data} and the \textit{land-cover predictions} derived through SCGCN using uncertainty simulation with DL\textsubscript{CycleGAN}, the full corpus of Straube maps is also presented. For illustrative reasons, aerial imagery from 1928 is included.
\\\\
\textbf{Implementation details}. 
All experiments, including the generation of synthetic training data using CycleGAN and UNSB, as well as domain-adaptive semantic segmentation on the generated historical maps with the adapted SCGCN, were performed using GPU acceleration via CUDA. The experiments were conducted in PyTorch 2.2.0 with CUDA 12.8 on an NVIDIA RTX 2000 Ada GPU equipped with 8 GB VRAM. 
Each bootstrapped training dataset comprised 2,269 image pairs, each containing a historical map and its corresponding annotated ground truth. For CycleGAN, we employed the official PyTorch implementation provided by \citet{cycleGAN}, setting the learning rate for both the generator and discriminator to $0.0002$, with a batch size of $1$. For UNSB, we used the PyTorch implementation from \citet{kim2023unsb}, also with a batch size of $1$, and added Gaussian noise to the input data with a standard deviation of $0.15$. For training the SCGCN, each dataset was randomly split into 80\% training and 20\% validation subsets. Importantly, the same random split was applied across both datasets to ensure fair and directly comparable results between models trained on the four different data variants. The SCGCN was trained using $32 \times 32$ nodes in the Self-Constructing Graph module, while inference was performed with $64 \times 64$ nodes. A batch size of $9$ and a learning rate of $3.47 \times 10^{-4}$ were used for all training runs.
\\\\
\textbf{FID}. The FID is a widely used metric for evaluating the quality of images generated with generative models by quantifying their similarity to real images \citep{paper:fid}. It measures the distance between two multivariate Gaussian distributions, representing the feature embeddings of real and generated images, using the Fréchet distance \citep{har2014frechet}. In our work, we use the FID to perform an unpaired image-to-image comparison between the generated historical maps and the Straube maps. Each distribution is parameterized by its mean $\mu$ and covariance matrix $\Sigma$, computed after transforming both sets of images from pixel space into a high-level feature space using the final pooling layer of an Inception-v3 network, as illustrated in Figure~\ref{architectureInception}.
\citep{Szegedy_2016_CVPR}. The FID is then computed, incorporating the trace $Tr$ of the covariance matrices as:
\begin{equation}\label{fid}
    FID = \left\|\mu_{X} - \mu_{Y}\right\|^{2} - Tr(\Sigma_{X} + \Sigma_{Y} - 2 \sqrt{\Sigma_{X}\Sigma_{Y}}).
\end{equation}
This enables the FID to capture perceptual differences based on semantic content rather than low-level pixel differences. Lower FID scores indicate greater similarity and therefore higher visual fidelity, while higher scores suggest larger discrepancies between the synthetic and real maps.
\begin{table}[ht!]
\centering
\small
\caption{FID scores for bootstrapped training datasets and Straube maps $\downarrow$.}
\begin{tabular}{ccccccc}
\toprule
\rowcolor{headergray}
dataset & historical-style & st. degraded & DL\textsubscript{CycleGAN} & DL\textsubscript{UNSB}\\ 
\midrule
Straube maps & 100.73 & 56.24 & 85.61~(Epoch~2) & \textbf{44.54}~(Epoch~400)\\
\bottomrule
\end{tabular}
\label{fid:table}
\end{table}
The FID results shown in Table~\ref{fid:table} indicate that the training dataset degraded by UNSB exhibits the highest similarity to the Straube maps. Interestingly, the purely historically-styled dataset shows the lowest similarity to the Straube maps compared to the stochastically degraded dataset or the datasets degraded by CycleGAN. We interpret this observation as preliminary evidence that the datasets with explicit simulation of aleatoric uncertainty are likely to achieve superior performance in semantic segmentation of the Straube maps, owing to the high similarity of their latent feature representations extracted from the final layer of the Inception-v3 network.

\subsection{Ablation Study}\label{ablation}

\begin{figure*}[ht!]
\centering
\subfloat[acc]{{\includegraphics[width=0.25\columnwidth]{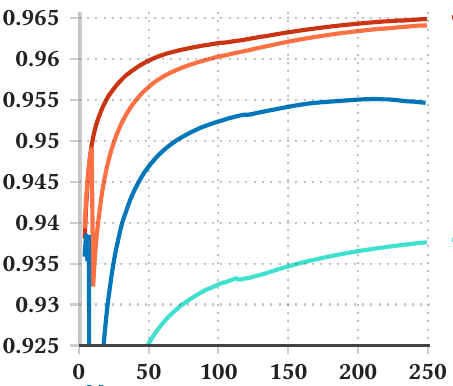}}\label{fig:acc}}
\hfill
\subfloat[acc cls]{{\includegraphics[width=0.25\columnwidth]{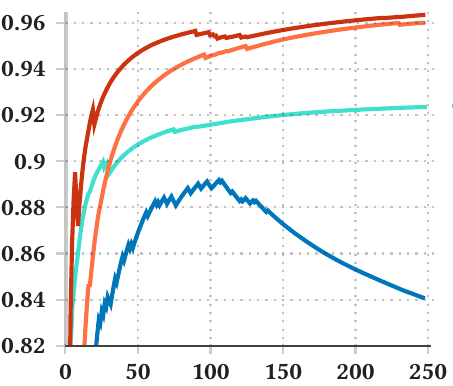}}\label{fig:acc_cls}}
\hfill
\subfloat[val loss]{{\includegraphics[width=0.25\columnwidth]{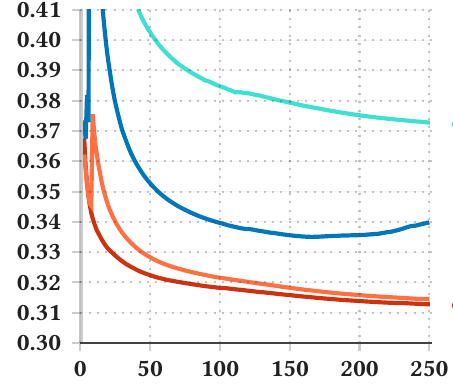}}\label{fig:val_loss}}
\hfill
\subfloat[main loss]{{\includegraphics[width=0.25\columnwidth]{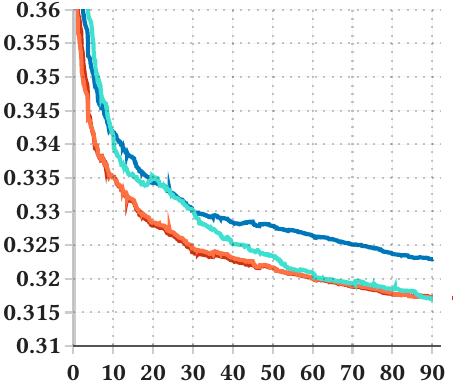}}\label{fig:main_loss}}
\hfill
\subfloat[mIoU]{{\includegraphics[width=0.25\columnwidth]{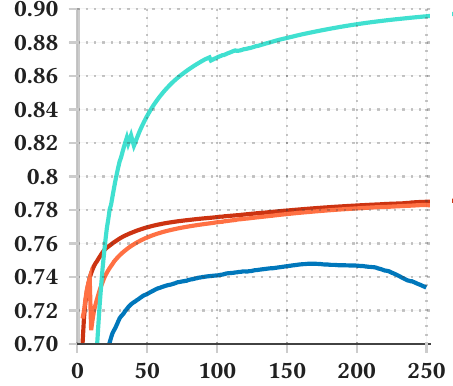}}\label{fig:miou}}
\hfill
\subfloat[Precision]{{\includegraphics[width=0.25\columnwidth]{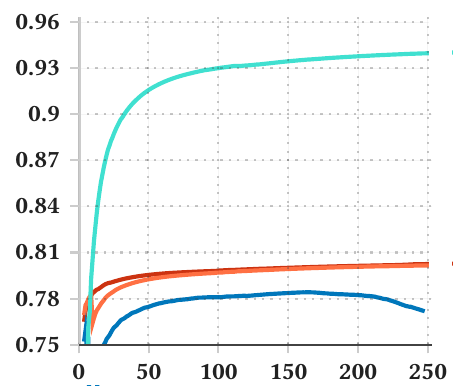}}\label{fig:precision}}
\hfill
\subfloat[Recall]{{\includegraphics[width=0.25\columnwidth]{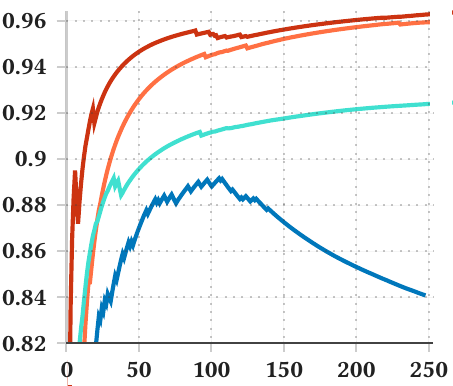}}\label{fig:recall}}
\hfill
\subfloat{{\includegraphics[width=0.25\columnwidth]{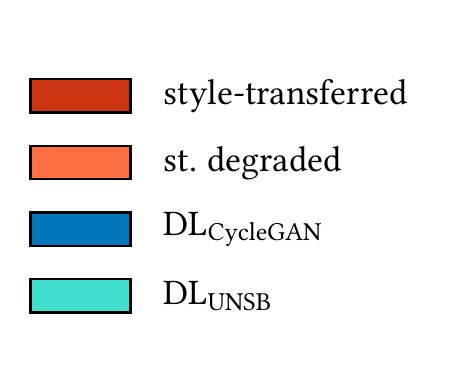}}}
\caption{Smoothed accuracy and loss curves during training (a-d) and validation metrics (e-g). The y-axis indicates the metric values, while the x-axis represents the number of epochs. For the main loss plot (d), the x-axis represents the number of training iterations, expressed in thousands.} 
\label{fig:training_val_metrics}
\end{figure*}
\begin{figure*}[ht!]
\centering
        \includegraphics[width=1\textwidth]{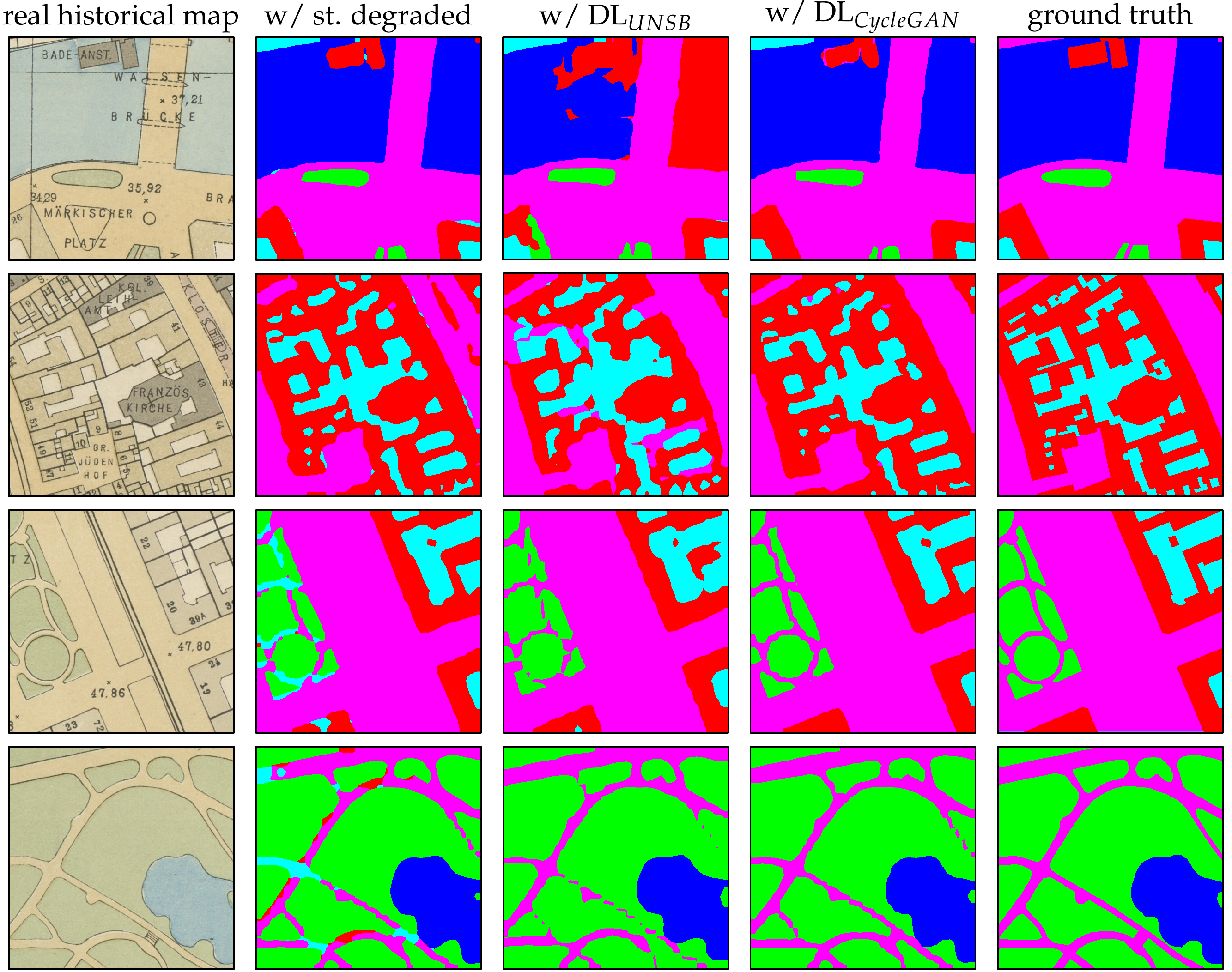}
        \label{fig:pairA}
    \caption{Four representative qualitative examples of domain-adaptive semantic segmentation obtained using bootstrapped training datasets with simulated aleatoric uncertainty. Blue denotes water bodies, turquoise presents sealed surfaces, red indicates buildings, magenta denotes infrastructure, and green represents recreational surfaces.}
    \label{fig:qualitative}
\end{figure*}

In the following, we evaluate and discuss the \textit{land-cover prediction} performance of the SCGCN models, which have been trained separately on each synthetically bootstrapped training dataset, on a homogeneous map corpus, namely the previously introduced Straube maps. To facilitate this evaluation and to individually assess the effectiveness of each training dataset, as mentioned, we manually annotated 125 Straube map patches, each of size 500×500 pixels, using the same five land-cover classes employed in generating the \textit{land-cover ground truth}, as detailed in Section~\ref{enum_classes}. 
\begin{table}[ht!]
\centering
\small
\setlength{\tabcolsep}{4.7pt}
\caption{Per-class and overall per-dataset percentage values of Precision, Recall, IoU and F$_{1}$ score, Accuracy and Kappa coefficient using the SCGCN after \cite{arzoumanidis:scgcn_2025} trained with bootstrapped datasets, evaluated on manually annotated test dataset.}
\begin{tabular}{>{\centering\arraybackslash}m{0.4cm} lcccccc}
\toprule
\rowcolor{headergray}
& \textbf{class} & Precision & Recall & IoU & F$_{1}$ score & Accuracy & Kappa\\
\midrule
\multirow{6}{*}{\rotatebox{90}{\shortstack{historically-style}}}
    & Buildings & 0.9578 & 0.6697 & 0.6506 & 0.7883 & 0.6697 & 0.6950\\
    & Infrastructure & \textbf{0.9437} & 0.9224 & \textbf{0.8743} & \textbf{0.9329} & 0.9224 & \textbf{0.9107} \\
    & Recr. surfaces & 0.9754 & 0.8913 & 0.8718 & 0.9315 & 0.8913 & 0.9216\\
    & Water bodies & 0.9553 & 0.9660 & 0.9243 & 0.9606 & 0.9660 & 0.9599\\
    & Sealed surfaces & 0.6152 & 0.9692 & 0.6034 & 0.7527 & 0.9692 & 0.6595\\
    & \textbf{Overall} & 0.7412 & 0.7364 & 0.6540 & 0.8732 & 0.8357 & 0.7796 \\
\midrule
\multirow{6}{*}{\rotatebox{90}{\shortstack{st. degraded}}}
    & Buildings & \textbf{0.9653} & 0.6109 & 0.5978 & 0.7483 & 0.6109 & 0.6464 \\ 
    & Infrastructure & 0.9340 & 0.8955 & 0.8423 & 0.9144 & 0.8955 & 0.8863\\
    & Recr. surfaces & \textbf{0.9788} & 0.9228 & 0.9047 & 0.9500 & 0.9228 & 0.9426\\
    & Water bodies & \textbf{0.9777} & 0.9718 & \textbf{0.9507} & \textbf{0.9747} & 0.9718 & \textbf{0.9742}\\
    & Sealed surfaces & 0.5762 & \textbf{0.9743} & 0.5676 & 0.7242 & \textbf{0.9743} & 0.6163\\
    & \textbf{Overall} & 0.7387 & 0.7292 & 0.6438 & 0.8623 & 0.8125 & 0.7499 \\
\midrule
\multirow{6}{*}{\rotatebox{90}{\shortstack{DL\textsubscript{CycleGAN}}}}
    & Buildings & 0.9555 & 0.7361 & \textbf{0.7117} & \textbf{0.8316} & 0.7361 & \textbf{0.7511}\\
    & Infrastructure & 0.8604 & \textbf{0.9583} & 0.8295 & 0.9068 & \textbf{0.9583} & 0.8730\\ 
    & Recr. surfaces & 0.9682 & 0.9354 & \textbf{0.9075} & \textbf{0.9515} & 0.9354 & \textbf{0.9442}\\
    & Water bodies & 0.9748 & \textbf{0.9732} & 0.9494 & 0.9740 & \textbf{0.9732} & 0.9736\\
    & Sealed surfaces & \textbf{0.7477} & 0.9520 & \textbf{0.7205} & \textbf{0.8375} & 0.9520 & \textbf{0.7832}\\
    & \textbf{Overall} & \textbf{0.7511} & \textbf{0.7592} & \textbf{0.6864} & \textbf{0.9003} & \textbf{0.8716} & \textbf{0.8268}\\
\midrule
\multirow{6}{*}{\rotatebox{90}{\shortstack{DL\textsubscript{UNSB}}}}
    & Buildings & 0.9107 & \textbf{0.7613} & 0.7085 & 0.8293 & \textbf{0.7613} & 0.7421\\
    & Infrastructure & 0.8739 & 0.9439 & 0.8309 & 0.9076 & 0.9439 & 0.8748\\
    & Recr. surfaces & 0.8823 & \textbf{0.9362} & 0.8323 & 0.9085 & \textbf{0.9362} & 0.8940\\
    & Water bodies & 0.9659 & 0.6555 & 0.6407 & 0.7810 & 0.6555 & 0.7775 \\
    & Sealed surfaces & 0.7460 & 0.8755 & 0.6744 & 0.8056 & 0.8755 & 0.7437\\
    & \textbf{Overall} & 0.7298 & 0.6954 & 0.6145 & 0.8464 & 0.8542 & 0.8024\\
\bottomrule
\end{tabular}
\label{tab:vertical-datasets}
\end{table}
This test dataset enables a robust evaluation of the domain-adaptive learning and model land-cover class prediction performance on an unseen historical map corpus. To ensure a fair and consistent comparison across all datasets, we used the same set of hyperparameters for training the SCGCN models for experiment. The experimental outcomes are assessed using quantitative metrics commonly used in machine learning tasks. As illustrated in Figure~\ref{fig:training_val_metrics}, the models demonstrate strong learning capabilities early in the training process, but with performance metrics only marginally improving over time. Due to the limited gains observed beyond this point, training was terminated after 250 epochs. 
\begin{figure}[ht!]
\centering
\subfloat[historically-styled]{{\includegraphics[width=0.25\columnwidth]{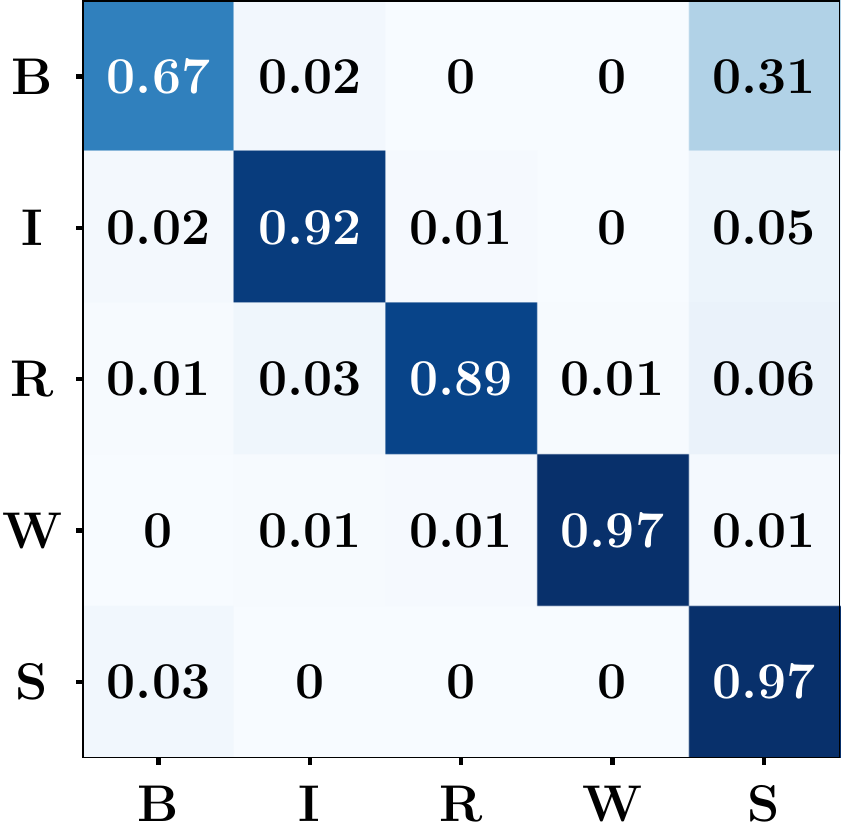}}\label{fig:conf_mat_handcrafted}}
\hfill
\subfloat[st. degraded]{{\includegraphics[width=0.25\columnwidth]{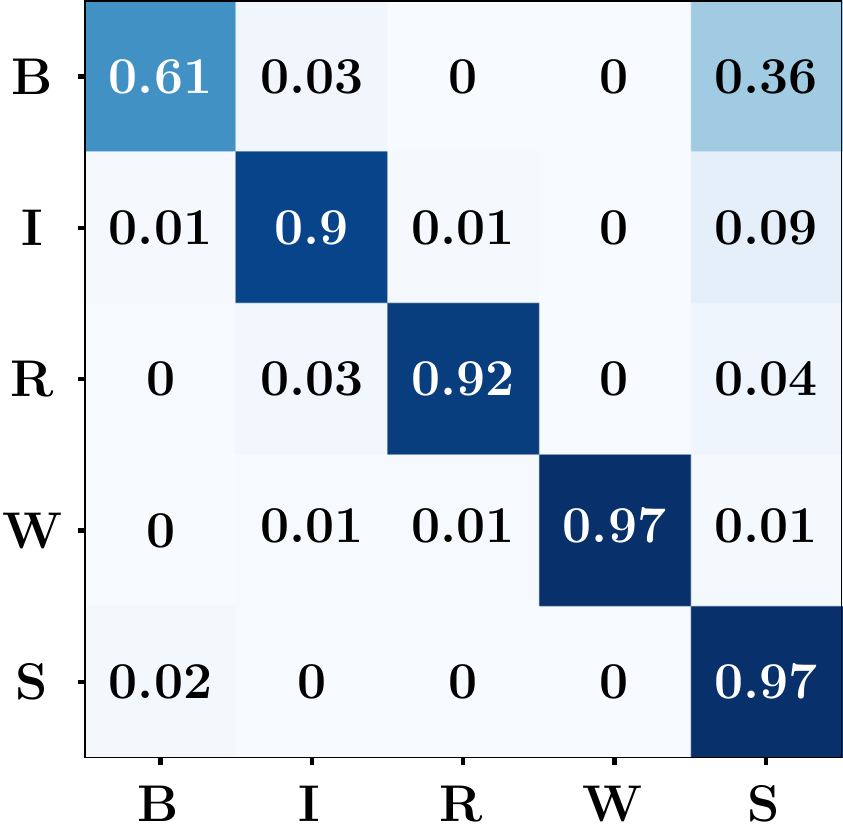}}\label{fig:conf_mat_enhanced_M}}
\hfill
\subfloat[DL\textsubscript{CycleGAN}]{{\includegraphics[width=0.25\columnwidth]{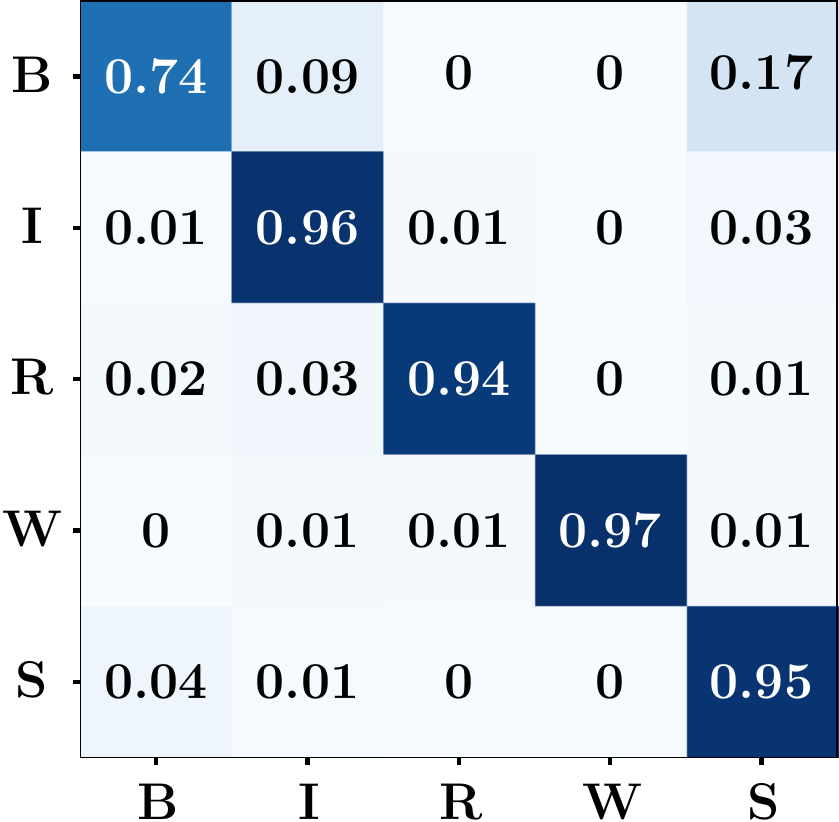}}\label{fig:conf_mat_enhanced_DL}}
\hfill
\subfloat[DL\textsubscript{UNSB}]{{\includegraphics[width=0.25\columnwidth]{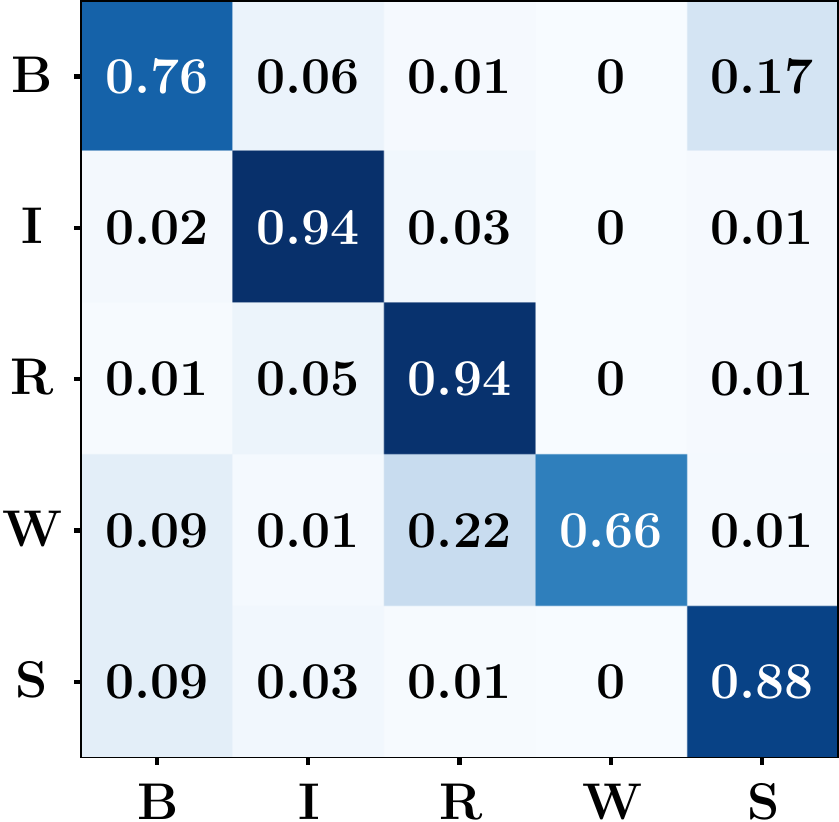}}\label{fig:conf_mat_enhanced_DL_stable}}
\caption{Normalized confusion matrices of land-cover class segmentation with accuracies rounded to two decimal places. The Y-axis represents the true labels, while the X-axis shows the predicted labels for all four different bootstrapped training datasets. The land-cover classes are abbreviated as follows: buildings (\textbf{B}), infrastructure (\textbf{I}), recreational surfaces (\textbf{R}), water bodies (\textbf{W}), and sealed surfaces (\textbf{S}).}
\label{fig:test}
\end{figure}
As shown in Figure~\ref{fig:training_val_metrics}, the SCGCN models trained on the purely historically-styled dataset and stochastically degraded synthetic dataset exhibit similar trends across all evaluation metrics. Correspondingly, the validation and main loss curves also suggest a more stable and effective learning process for these models when compared to the models trained on automatically degraded synthetic datasets. Notably, the DL\textsubscript{CycleGAN} dataset exhibits the weakest performance across nearly all metrics during training. Since only marginal performance improvements were observed beyond epoch 100, this epoch was selected for the semantic segmentation evaluation. 
Interestingly, evaluation of the model performance on the test dataset indicates that the DL\textsubscript{CycleGAN} dataset achieves the highest scores across all evaluation metrics, as presented in Table~\ref{tab:vertical-datasets}. With respect to accuracy and the Kappa coefficient, the model trained on the DL\textsubscript{CycleGAN} dataset performs approximately 4\% better than the historically-styled dataset, 6–7\% better than the stochastically degraded dataset, and 2\% better than the DL\textsubscript{UNSB} dataset. However, it is worth noting that although the DL\textsubscript{CycleGAN} dataset slightly outperforms the DL\textsubscript{UNSB} dataset, both automatically degraded datasets are competitive, with the GAN-based dataset showing a consistent but modest advantage. Based on this observation, we hypothesize that prolonged training on bootstrapped datasets can diminish the model’s generalization capability to unseen \textit{real historical maps}. This limitation is likely attributable to overfitting on the synthetic training distribution, especially when the bootstrapped datasets exhibit insufficient realism and variation in simulated aleatoric uncertainty. An analysis of the normalized confusion matrices in Figure~\ref{fig:test} indicates that, across all datasets, the highest rate of misclassification occurs in the buildings land-cover class, which is most frequently confused with the sealed surface class. The only exception is the model trained on the DL\textsubscript{UNSB} dataset, where the highest misclassification rate is observed in the water class, predominantly confused with recreational surfaces. Notably, the model with the highest metric scores overall trained on the DL\textsubscript{CycleGAN} dataset misclassifies buildings as infrastructure (approximately~9\%) and with sealed surfaces~(17\%). This observation is further supported by the fact that the same model appears to visually perform best in accurately identifying infrastructure elements, such as streets, as can be identified in Figure~\ref{fig:qualitative}. 
\\\\
Interestingly, the IoU and F$_1$ scores for the land-cover classes buildings, recreational surfaces, and sealed surfaces are highest for the model trained on the DL\textsubscript{CycleGAN} dataset, as shown in Table~\ref{tab:vertical-datasets}, a pattern that is also evident in the qualitative results in Figure~\ref{fig:qualitative}. This finding is noteworthy because the individual recall and precision values for these classes are slightly lower than those of the respective best-performing models, except for the precision of sealed surfaces, where the model trained on the DL\textsubscript{CycleGAN} dataset marginally outperforms all others. Although the model trained on the DL\textsubscript{CycleGAN} dataset achieves the highest accuracy (cf. Figure~\ref{fig:conf_mat_enhanced_DL}) and recall (cf. Table~\ref{tab:vertical-datasets}) for the infrastructure class, it exhibits a markedly lower IoU~(-5\%) and F$_1$ score~(-3\%) compared to the best-performing model in this case, which is the purely historically-styled dataset. This discrepancy suggests that the model may be overly liberal in predicting infrastructure, potentially leading to over-segmentation. Given that infrastructure often lies adjacent to both buildings and sealed surfaces, this could explain the reduced precision and overall performance for this class.
\\\\
Visually, the datasets with automatic degradation used for semantic segmentation produce results with higher perceptual fidelity which is also supported by the quantitative evaluation on the test dataset. This may be due to the observation that deep generative approaches introducing excessive aleatoric uncertainty, also present in the test dataset. Although often imperceptible to the human eye, this noise can positively impact the semantic segmentation model's performance, ultimately leading to improved accuracy. 
\\\\
Although diffusion-based image generation methods such as UNSB generally produce images that are perceptually closer to the ground-truth image distribution than GAN-based approaches \citep{NEURIPS2021_49ad23d1, NEURIPS2020_4c5bcfec}, as also reflected in Table~\ref{fid:table}, they do not yield the most accurate results for the semantic segmentation of Straube maps when used to bootstrap a training dataset to train a SCGCN model. As described in Section~\ref{data_uncertainty}, the operating characteristics of CycleGAN allow it to be stopped early, thereby preserving semantic consistency with the input map, whereas results generated by UNSB do not retain this consistency. This property appears to be crucial for the performance of a trained SCGCN model in the semantic segmentation task, as the generated historical maps are not guaranteed to be perfectly aligned with the corresponding \textit{land-cover ground truth}, even though such alignment is important for supervised deep learning models such as SCGCN.
\\\\
Although our approach was evaluated on the downstream task of semantic segmentation, it may also be advantageous for other applications, such as image restoration and object detection. We chose semantic segmentation as the primary evaluation task because it requires dense, pixel-level classification and is therefore particularly well suited to assessing whether the generated training data faithfully capture the visual and structural characteristics of historical maps. In this sense, semantic segmentation provides a rigorous and representative benchmark for evaluating the utility of our approach in downstream analysis.
\\\\
A particularly promising application area is the detection and classification of road lines in historical maps. As discussed in Section~\ref{sec:related_work}, existing approaches that rely on historically styled training data generally require georeferenced historical vector data that correspond precisely to the spatial layout of the historical map under analysis, in this case for road detection and classification \citep{jiao:roadVector, JIAO2022102980}. In contrast, our approach offers the advantage that it can be applied without requiring historical prior knowledge or georeferenced data beyond map-intrinsic characteristics, such as colours, line thicknesses, hachures, and common symbols. This substantially increases the practical feasibility of applying our method to downstream tasks of this kind.
\\\\
In addition, our approach may also be applicable to map restoration. As discussed earlier, existing methods for the restoration of historical maps typically rely on georeferenced map pairs for training in order to perform paired image generation, although such data are scarce in practice \citep{affolter_generative_2025}. Since our approach does not depend on pairwise georeferenced examples of modern and historical map pairs, it could also be used for map restoration, for example to artificially reconstruct damaged areas of historical maps.

\section{Conclusion and Future Work}\label{conclu}
Developing an approach capable of automatically generating \textit{unlimited} quantities of annotated historical maps represents a crucial step toward the comprehensive digitization and interpretation of historical cartographic sources. This is particularly important because most state-of-the-art supervised methods for automatic map interpretation rely on deep learning, which inherently demands large volumes of annotated training data. In this work, we introduced an automated framework that transfers the cartographic style of specific historical map corpora onto modern vector data, such as OpenStreetMap (OSM). To enhance the realism and domain fidelity of the synthesized historical maps, our approach automatically applies corpus-specific degradations that simulate aleatoric uncertainties, including noise originating material deterioration and scanning artifacts. We assessed the effectiveness of the proposed framework through a comprehensive evaluation using a deep learning-based semantic segmentation model trained exclusively on the generated synthetic maps. The results demonstrate that domain-adaptive semantic segmentation trained solely on synthetically bootstrapped data can achieve an accuracy of approximately 88\% when applied to \textit{real historical maps}, underscoring the potential of our method for large-scale, automated map interpretation. 
\\\\
Although \citet{hurni:spatioTemporal} argue that increasing the amount of training data alone cannot resolve aleatoric uncertainty, our results demonstrate that simulating such uncertainties within synthetically generated training data can lead to highly successful deep learning-based semantic segmentation of homogeneous historical map corpora even in the absence of any real ground truth data. By reducing the time required to generate annotated training data to just a few hours, our approach significantly lowers the cost and effort involved.
\\\\
Future work will focus on improving the geometric and semantic consistency of the generated maps by incorporating automatic mechanisms that explicitly enforce spatial coherence and simulate uncertainty in a more principled manner. One possible direction for future work is the adoption of loss functions designed to enforce geometric consistency during generation, thereby enabling GAN models to be trained over longer epochs without loss of semantic consistency and spatial coherence. Another promising avenue involves integrating the proposed framework with advanced diffusion-based generative models, such as Stable Diffusion (SD), in combination with ControlNet~\citep{Zhang_2023_ICCV} to further enhance style transfer by constraining the generation process through explicit semantic layout conditioning. Ultimately, this combination could eliminate the need for manual style transfer and fully automate the generation of high-fidelity, semantically aligned synthetic historical maps.
\\\\
This work represents an important milestone toward unlocking the vast potential of digitized historical map archives. By enabling the automatic and scalable generation of annotated cartographic data, our approach paves the way for the large-scale interpretation of thousands of scanned historical maps, revealing spatial, cultural, and environmental information that has remained inaccessible to computational analysis for centuries.

\section*{Acknowledgements}
We gratefully acknowledge the Staatsbibliothek zu Berlin for providing access to the \textit{Julius Straube} map collection. We further extend our thanks to Jannik Matijevice for his assistance in annotating the test data employed for evaluation and Lena Hildebrandt for developing the web viewer used to visualize the generated datasets and segmentation results.

\bibliography{sn-bibliography}


\end{document}